\documentclass{article}

 \usepackage[dblblindworkshop, final]{neurips_2025}
\workshoptitle{Mechanistic Interpretability}

\usepackage{amsmath}
\usepackage{amssymb}
\usepackage{mathtools}
\usepackage{placeins}

\usepackage[utf8]{inputenc} %
\usepackage[T1]{fontenc}    %
\usepackage{hyperref}       %
\usepackage{url}            %
\usepackage{booktabs}       %
\usepackage{amsfonts}       %
\usepackage{nicefrac}       %
\usepackage{microtype}      %
\usepackage{xcolor}         %
\usepackage{natbib}         %
\title{Compressed Computation is (probably) not Computation in Superposition}

\author{%
  Jai Bhagat$^*$ \\
  Metamorphic \\
  \And
  Sara Molas-Medina$^*$ \\
  Independent \\
  \And
  Giorgi Giglemiani \\
  UK AI Security Institute \\
  \And
  Stefan Heimersheim\footnotemark[2] \\
}

\begin{document}

\maketitle

\begin{abstract}
  We study whether the Compressed Computation (CC) toy model \citet{braun2025interpretability} is an instance of computation in superposition. The CC model appears to compute 100 ReLU functions with just 50 neurons, achieving a better loss than expected from only representing 50 ReLU functions. We show that the model mixes inputs via its noisy residual stream, corresponding to an unintended mixing matrix in the labels. Splitting the training objective into the ReLU term and the mixing term, we find that performance gains scale with the magnitude of the mixing matrix and vanish when the matrix is removed. The learned neuron directions concentrate in the subspace associated with the top 50 eigenvalues of the mixing matrix, suggesting that the mixing term governs the solution. Finally, a semi-non-negative matrix factorization (SNMF) baseline derived solely from the mixing matrix reproduces the qualitative loss profile and improves on prior baselines, though it does not match the trained model. These results suggest CC is not a suitable toy model of computation in superposition.
\end{abstract}

\renewcommand{\thefootnote}{\fnsymbol{footnote}} %
\footnotetext[1]{These authors contributed equally to this work and are listed in alphabetical order.} %
\footnotetext[2]{Correspondence to \texttt{stefan.heimersheim@gmail.com}. Work done while at Apollo Research.}
\renewcommand{\thefootnote}{\arabic{footnote}}
\setcounter{footnote}{0}

\section{Introduction}
Superposition lets networks represent many sparse features in fewer dimensions \citep{elhage2022toy, gurnee2023finding, bricken2023towards}. The related question, \emph{computation in superposition}, asks whether models can implement more nonlinear functions than they have nonlinearities when inputs are sparse \citep{hanni2024mathematical, bushnaq2024circuits, adler2024complexity, olah2025toy}. 
\citet{braun2025interpretability} introduce a toy model of Compressed Computation (CC)
which seemingly implements computation in superposition: it appears to
compute 100 ReLU functions of sparse inputs using only 50 neurons.

We re-examine the CC model, and find that its performance---a better
loss than a naive baseline computing only 50 ReLU functions---is likely due to
noisy labels introduced by \citet{braun2025interpretability}'s architecture.
Our key results are:
\begin{itemize}
  \item The CC model's performance advantage is dependent on residual stream noise that mixes different inputs into the labels (in addition to the ReLU target). The model does not beat baselines without this effective ``mixing matrix''.
  \item The CC model's performance scales smoothly with the mixing matrix's magnitude (higher is better, up to a limit). We also find that the trained model focuses on the top 50 singular vectors of the mixing matrix (all its neuron directions mostly fall into this subspace), mostly ignoring the 50 others.
  \item We introduce a new baseline model derived from the SNMF of the mixing matrix alone which beats the previous naive baseline, and qualitatively matches the CC model's loss profile. However, the SNMF weights qualitatively differ from the trained model's weights, and we don't fully recover the trained model's loss.
\end{itemize}

\section{Methods}
We show the residual CC architecture with fixed random embedding
$W_E$ \citep{braun2025interpretability} is equivalent to a 1-layer
MLP trained on $y=\mathrm{ReLU}(x)+Mx$ with
$M=I-W_EW_E^T$ (details in appendix \ref{app:cc-equivalence}). Inputs
$x_{0..99}$ are nonzero with probability $p$ and drawn uniformly from
$[-1,1]$, otherwise zero. We vary $M$ across three conditions:
$M=0$ (clean), random $M\sim \mathcal{N}(0, \sigma^2)$, and the embedding-induced
$M=I-W_EW_E^T$. Only $W_{\text{in}}$ and $W_{\text{out}}$ are
trained. Training details are provided in appendix \ref{app:training}.
We provide our core code and scripts to reproduce all figures at
\href{https://osf.io/dhqf7/files/osfstorage?view_only=7a036c7c8da342e6bb72f08cc7352c71}{this URL}.

\begin{figure}[ht]
  \centering
  \includegraphics[width=1\linewidth]{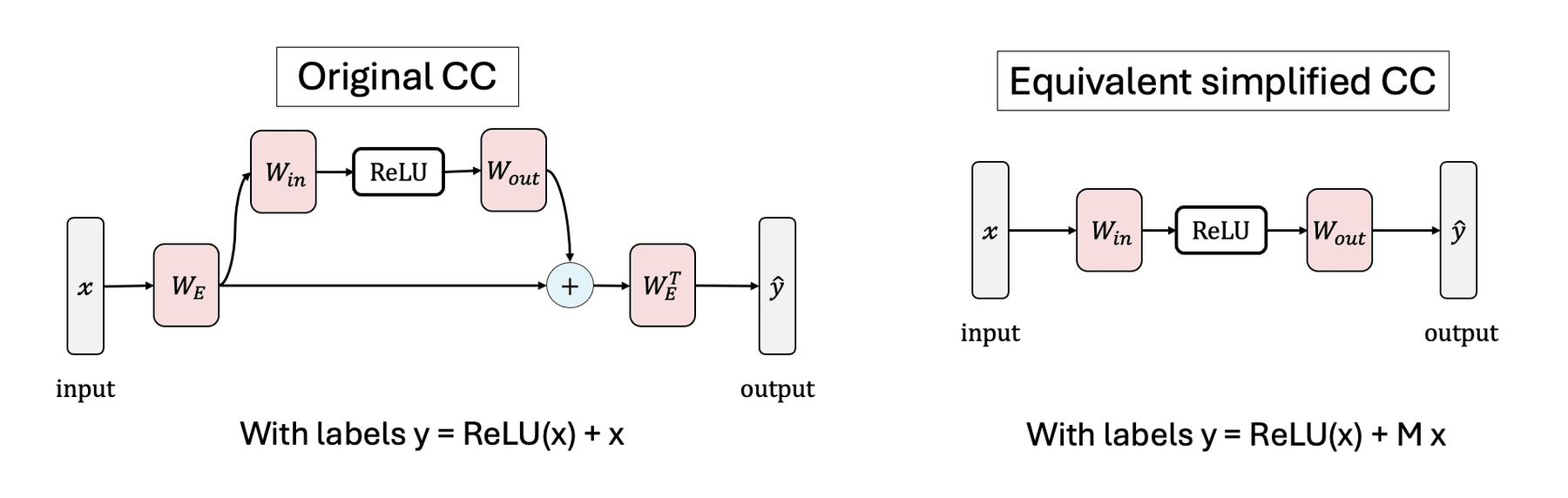}
  \caption{The original model architecture from \citet{braun2025interpretability}, and our simpler equivalent model. The labels for our (new) model are $y_i = \text{ReLU}(x_i) + \sum_j M_{ij}x_i$. The matrix $M$ mixes other inputs $x_j$ into the label $y_i$. Thus the MLP needs to learn both the ReLU term, and the mixing term.}
  \label{fig1:architecture}
\end{figure}

\section{Results}
\paragraph{Simplified model reproduces CC results.} We reproduce the results of
\citet{braun2025interpretability} with our simplified model. Our
1-layer MLP model is trained on $y = \text{ReLU}(x) + Mx$ with a mixing matrix
$M = 1 - W_E W_E^T$. This mixing matrix accounts for the noisy residual stream
contribution in \citet{braun2025interpretability}'s model. Our model
produces the same results, and we reproduce their figures in
appendix \ref{app:cc-analysis}.
Figure \ref{fig5:loss} (left) shows that the ``better-than-naive\footnote{The
naive baseline of representing 50 ReLU functions perfectly achieves a loss of
0.0833 on a clean dataset (details in appendix \ref{app:naive}).}
at low $p$ / worse at high $p$''
profile is reproduced by the embedding-like $M$, as well as by a random
$M\sim \mathcal{N}(0, 0.02)$. We find that a symmetric random $M$ matrix matches
\citet{braun2025interpretability}'s results almost exactly.

\begin{figure}[h]
  \centering
  \includegraphics[width=0.8\linewidth]{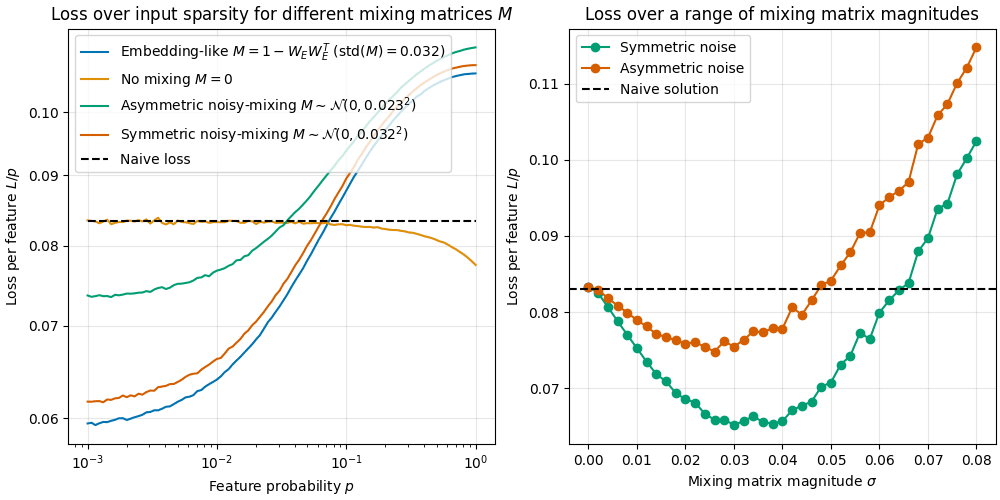}
  \caption{Left: With $M\neq0$ (embed-like, random, or symmetric),
  the model beats the naive loss at high sparsity. Labels with
  embed-like $M$ allow for the lowest loss, but random $M$ matrices
  behave similarly.
  Right: Optimal loss decreases with noise scale $\sigma$ for small $\sigma$ and
  increases again at large $\sigma$.}
\label{fig5:loss}
\end{figure}

\begin{figure}[h]
\centering
\includegraphics[width=0.8\linewidth]{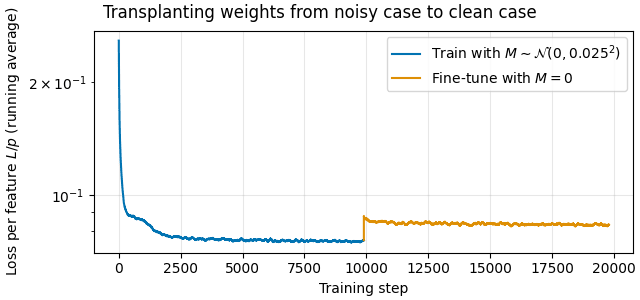}
\caption{Transplant to clean labels. A model trained with
$M\neq0$ immediately reverts to the naive loss when fine-tuned on $M=0$, indicating the lower loss is not just due to training dynamics.}
\label{fig6:transplant}
\end{figure}

\paragraph{CC loss advantage requires a nonzero mixing matrix.} Without the noise-induced mixing matrix, the model does not beat the naive loss (yellow line in Figure \ref{fig5:loss} left) for sparse
inputs.\footnote{The dense regime is a different mechanism, discussed in appendix \ref{app:dense-analysis}.}
If the model was achieving a good loss due to computation in superposition, we would expect the loss
to still beat the naive loss even with $M=0$. Furthermore, we confirm that a model trained on $M\neq0$
immediately reverts to the naive loss when fine-tuned on an $M=0$ dataset (Figure \ref{fig6:transplant}).

\paragraph{CC loss advantage scales with the magnitude of $M$.} We train CC models
over a range of noise scales $M\sim \mathcal{N}(0, \sigma)$ with $\sigma$ from 0 to
0.08, as shown in Figure \ref{fig5:loss} (right). We find that the loss
decreases with $\sigma$ (for $\sigma \lesssim 0.03$). This direct correspondence
strongly suggests that the loss advantage is coming directly from the noise term,
rather than from computing the 100 ReLU target functions (computation in superposition).

\begin{figure}[h]
\centering
\includegraphics[width=0.8\linewidth]{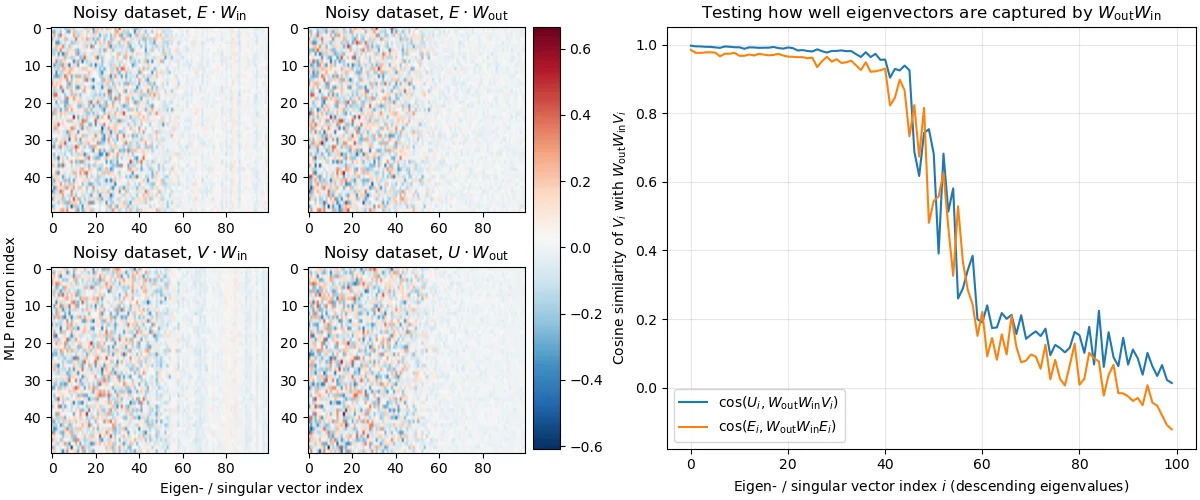}
\caption{Left: Neuron read/write directions align with the top
$\sim50$ eigen/singular vectors of $M$. Right: The linear map
$W_{\rm out}W_{\rm in}$ preserves those top directions while
strongly attenuating the rest.}
\label{fig7:svd}
\end{figure}

\paragraph{Potential mechanism behind the Compressed Computation model.}
We notice that the direction read ($W_{\rm in}$) and written ($W_{\rm out}$) by each
MLP neuron lies mostly in the 50-dimensional subspace spanned by the positive eigenvalues
of $M$ (Figure \ref{fig7:svd} left, top panels) or the top 50 singular vectors of $I+M$ (bottom panels).
We confirm this by measuring the cosine similarity of singular- and eigenvectors with their respective projections through $(W_{\rm out}W_{\rm in})$. We find that only the top $\sim 50$ vectors are represented fully, with a sharp drop-off in similarity around index 50 (Figure \ref{fig7:svd} right).

\begin{figure}[h]
\centering
\includegraphics[width=0.8\linewidth]{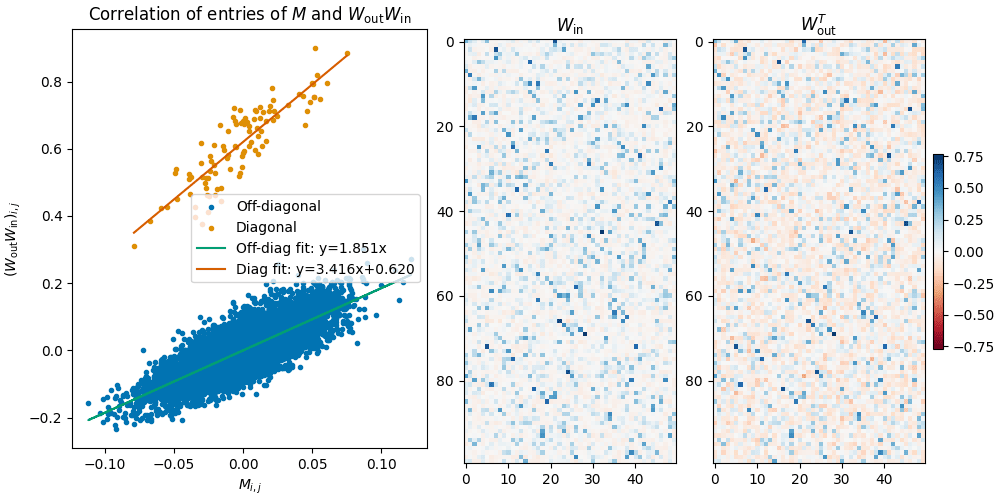}
\caption{Left: Entries of the $M$ matrix strongly correlated with entries of $W_{\rm out}W_{\rm in}$ (with an offset for the diagonal entries). Right: $W_{\rm in}$ is mostly nonnegative and both weight matrices are sparse.}
\label{fig8:weights}
\end{figure}

\begin{figure}[h]
\centering
    \includegraphics[width=0.7\linewidth]{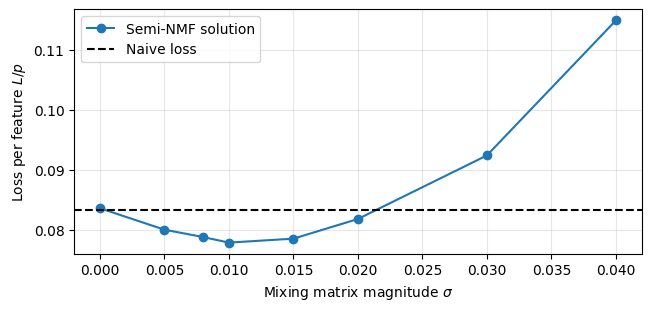}
\caption{The semi-NMF baseline derived from $M$ reproduces the
qualitative loss-vs-$\sigma$ curve, but does not reach as low a loss as the trained model.}
\label{fig9:snmf}
\end{figure}

We further analyze the trained model's weights and find (i) the entries of the $M$ matrix are strongly correlated with the entries of $W_{\rm out}W_{\rm in}$ (Figure \ref{fig8:weights} left), and (ii) $W_{\rm in}$ is mostly non-negative and both weight matrices are sparse (Figure \ref{fig8:weights} right).
Given this we attempt to design hand-coded model weights using the SNMF of $M$, setting $W_{\rm in}$ to the non-negative factor. We find that this SNMF solution beats the naive loss for a range of noise
scales $\sigma$ (Figure \ref{fig9:snmf}), in a qualitatively similar pattern to the
trained model (Figure \ref{fig5:loss} right). However, the SNMF solution does not reach
a loss as low as the trained model, and its weights are less sparse than the trained
weights (Figure \ref{fig8:weights} right).

\section{Conclusion}
Our work sheds light on the mechanism behind the \citet{braun2025interpretability} model of compressed
computation. We ruled out the hypothesis that the embedding noise was only required for
the training of the model, and instead showed that the noise-induced mixing matrix is
what allows the model to beat the naive baseline.
We provided an improved SNMF baseline that explains some but not all of the trained
model's performance and properties.
While we were not able to fully reverse-engineer the Compressed Computation model,
we hope that our work inspires new attempts to find a toy model of computation in
superposition, and further reverse-engineer the CC model.

\bibliography{references}
\appendix

\section{Mathematical details}
\subsection{Equivalence of the CC model and 1-layer MLP}
\label{app:cc-equivalence}
The CC residual architecture corresponds to the function
\begin{align}
  \hat y' &= W_E^T \left( W_E x + W'_{\rm out} \mathrm{ReLU}(W'_{\rm in} W_E x) \right) \\
  &= \underbrace{W_E^T W_E}_{I - M} x + \underbrace{W_E^T W'_{\rm out} \mathrm{ReLU}(W'_{\rm in} W_E x)}_{W_{\rm out} \mathrm{ReLU}(W_{\rm in} x)}
\end{align}
and is trained on the objective $y' = \mathrm{ReLU}(x) + x$. This is equivalent
to training a 1-layer MLP
\begin{align}
  \hat y &= W_{\rm out} \mathrm{ReLU}(W_{\rm in} x)
\end{align}
on the objective $y = \mathrm{ReLU}(x) + Mx$.

\subsection{Naive baseline loss}
\label{app:naive}
A noise-free model could implement 50 ReLU target functions perfectly,
and ignore the other 50 target functions. The expected loss for an
unrepresented feature is then $0.5 \int_{-1}^1 \text{ReLU}(x)^2 \mathrm{d}x \approx 0.1667$.
Thus the average loss per feature is 0.0833 for the naive baseline.

This only holds on the clean dataset ($M=0$). With noisy labels, the naive solution
would need to be adjusted to account for the noise. We do not investigate this further,
and only indicate the 0.0833 line for comparison with \citet{braun2025interpretability}.

\section{Training details}
\label{app:training}
We use a batch size of 2048 and a learning rate of 0.003 with cosine
scheduler. To control for training exposure we train all models for
10,000 non-empty batches. The only trainable parameters are $W_{\rm in}$
and $W_{\rm out}$ (not $M$). In some experiments we optimize a scale factor
$\sigma$ for the mixing matrix $M$ simultaneously with the weights for
convenience.

\section{Analysis of the CC model}
\label{app:cc-analysis}
We reproduce the CC model of \citet{braun2025interpretability}
with our simplified architecture with $M = 1 - W_E W_E^T$ for a randomly
generated $\mathbb{R}^{1000\times 100}$ embedding matrix $W_E$, as well
as for a random matrix $M \sim \mathcal{N}(0, 0.02)$.

\begin{figure}[h]
  \centering
  \includegraphics[width=0.8\linewidth]{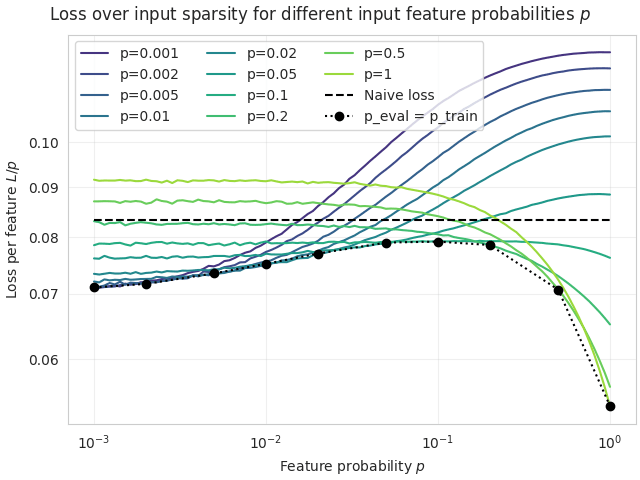}
\caption{Loss per feature ($L/p$) as a function of evaluation sparsity. Each solid line corresponds to a model trained at a given sparsity. The models learn one of two solution types, depending on the input sparsity used during training: the CC solution (violet) or a dense solution (green). Both types beat the naive baseline (dashed line) in their respective regime. Black circles connected by a dotted line represent the results seen by \citet{braun2025interpretability}, where models were evaluated only at their training sparsity.}
\label{fig2:loss}
\end{figure}

In Figure \ref{fig2:loss} we show the loss per feature (that is, MSE loss
divided by feature probability $p$) as a function of training and
evaluation sparsity.

\begin{figure}[h]
  \centering
      \includegraphics[width=0.8\linewidth]{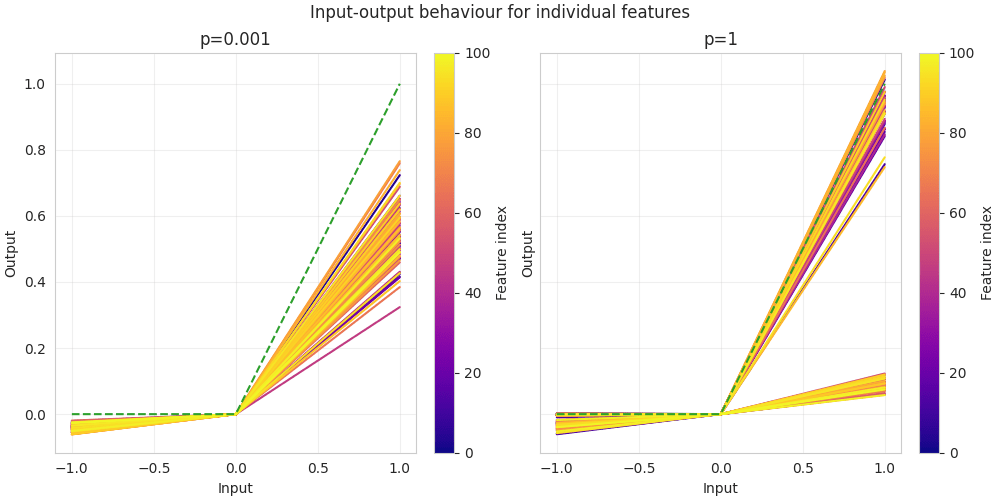}
  
  \caption{Input/output behaviour of the two model types (for one-hot inputs): In the CC solution (left panel), all features (inputs) are similarly-well represented: each input activates the corresponding output feature. In contrast, the dense solution (right panel) shows a strong (and more accurate) response for half the features, while barely responding to the other half. The green dashed line indicates the expected response under perfect performance.}
  \label{fig3:input-output}
  \end{figure}
  
  \begin{figure}[h]
  \centering
  \includegraphics[width=0.8\linewidth]{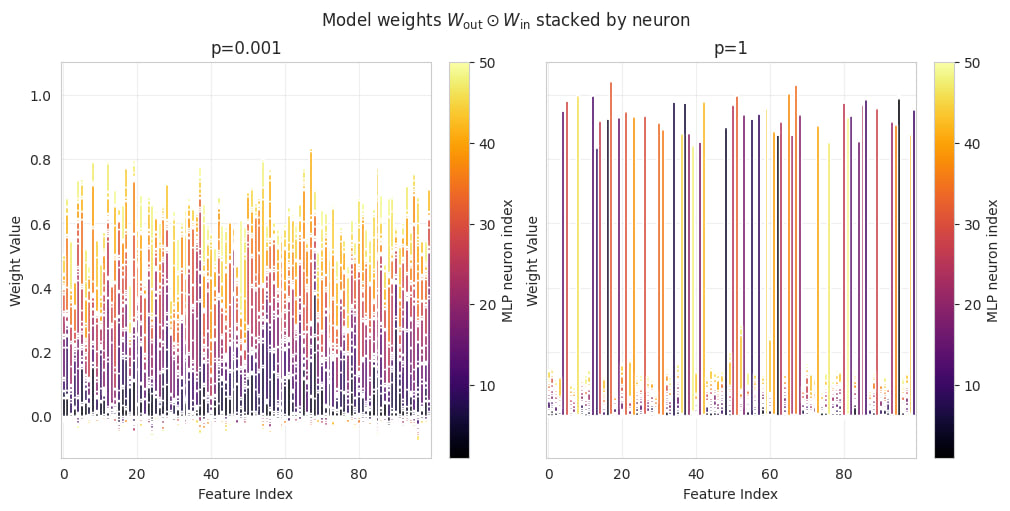}
  \caption{Weights representing each input feature, split by neuron. Each bar corresponds to a feature (x-axis) and shows the adjusted weight value from $W_{\rm out} \odot W_{\rm in}$, split by neuron index (color). The CC solution (left) combinations of neurons to represent each feature (to around 70\%), whereas the dense solution (right) allocates a single neuron to fully ($\sim$100\%) represent 50 out of 100 features.}
  \label{fig4:weights}
  \end{figure}

In the sparse regime (low probability $p \lesssim 0.05$) we find solutions that perform
well on sparse inputs, and less well on dense inputs. They typically exhibit a similar
input-output response for all features (Figure \ref{fig3:input-output} left), and weights distributed across all
features (Figure \ref{fig4:weights} left, equivalent to Figure 6 in \citet{braun2025interpretability}). The maximally sparse
case (exactly one input active) behaves very similar to $p \leq 0.01$.

In the dense regime (high probability $p \gtrsim 0.2$) we find solutions
with a constant per-feature loss on sparse inputs, but a better performance
on dense inputs. These solutions tend to implement half the input features
with a single neuron each, while ignoring the other half (Figures
\ref{fig3:input-output} right and \ref{fig4:weights} right).

\FloatBarrier
\section{Mechanism of the dense solution}
\label{app:dense-analysis}
The dense regime is less interesting as it is not related to computation in superposition,
however Figure \ref{fig2:loss} shows that the model can beat the naive baseline in the dense regime.
In this section we provide an explanation of this behaviour, and a hand-coded model that matches
the trained model's performance in the dense regime.

Revisiting Figures \ref{fig3:input-output} and \ref{fig4:weights}, we notice an intriguing pattern:
roughly half of the features are well-learned, while the other half are only weakly represented.
Our hypothesis is that the model represents half the features correctly, and approximates the other half by emulating a bias term. Our architecture does not include biases, but we think the model can create an offset in the outputs by setting the corresponding output weight rows to positive values, averaging over all features. This essentially uses the other features to, on average, create an offset.

\begin{figure}[h]
  \centering
  \includegraphics[width=0.8\linewidth]{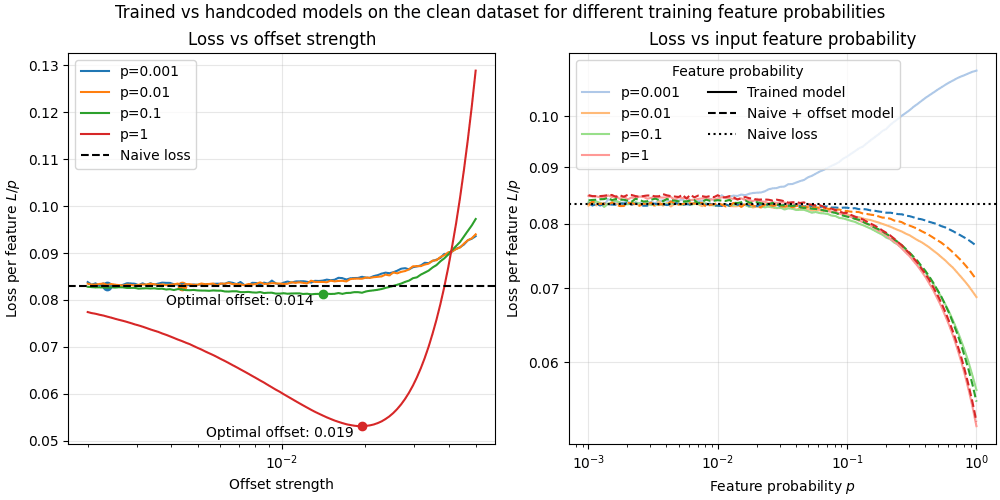}
\caption{Left: A non-zero offset in the $W_{\rm out}$ entries of unrepresented features improves the loss in the dense regime. We determine the optimal value empirically for each input feature probability $p$.
Right: This hand-coded naive + offset model (dashed lines) consistently matches or outperforms the model trained on clean labels (solid lines) in the dense regime. (Note that this plot only shows the clean dataset ($M = 0$) which is why no solution outperforms the naive loss in the sparse regime.)}
\label{fig10:dense-analysis}
\end{figure}

We test this ``naive + offset'' solution on the clean dataset ($M = 0$) as the behaviour seems to be the same regardless of noise (but we did not explore this further), and find that the hardcoded naive + offset solution does in fact match the trained models' losses. Figure \ref{fig10:dense-analysis} left shows the optimal weight value (offset strength; same value for every $W_{\rm out}$ entry of non-represented features), and Figure \ref{fig10:dense-analysis} right shows the corresponding loss as a function of feature probability. We find that the hardcoded models (dashed lines) closely match or exceed the trained models (solid lines).
We thus conclude that the high density behaviour can be explained by a simple bias term, and not particularly interesting.

\end{document}